\documentclass[AMA,STIX1COL]{WileyNJD-v2}

\newcommand\BibTeX{{\rmfamily B\kern-.05em \textsc{i\kern-.025em b}\kern-.08em
T\kern-.1667em\lower.7ex\hbox{E}\kern-.125emX}}

\articletype{Research article}%

\usepackage{amsmath}

\usepackage[ruled,linesnumbered,algo2e]{algorithm2e}

\usepackage{latexsym}
\usepackage{bm,bbm}
\usepackage{centernot}
\usepackage{ragged2e}
\usepackage{moreverb}
\usepackage{xcolor}
\usepackage{command}
\usepackage{graphicx}
\usepackage{mathtools}
\usepackage{float}
\usepackage{comment}
\usepackage{subcaption}
\usepackage{wrapfig}
\usepackage{lipsum}
\usepackage{xcolor}

\begin{document}

\title{DeepPhysics: a physics aware deep learning framework for real-time simulation}
\author[1]{Alban Odot}
\author[1]{Ryadh Haferssas}
\author[1]{Stephane Cotin}

\authormark{DeepPhysics Odot Haferssas Cotin}

\address[1]{\orgdiv{MIMESIS team}, \orgname{Inria}, \orgaddress{1 Place de l'Hopital, 67000 Strasbourg, \country{France}}}

\corres{Stephane Cotin\\\email{stephane.cotin@inria.fr}}

\abstract[Abstract]
{Real-time simulation of elastic structures is essential in many applications, from computer-guided surgical interventions to interactive design in mechanical engineering. The Finite Element Method is often used as the numerical method of reference for solving the partial differential equations associated with these problems. Yet, deep learning methods have recently shown that they could represent an alternative strategy to solve physics-based problems \cite{raissi2019physics,sanchez2020learning,kim19deep}.
In this paper, we propose a solution to simulate hyper-elastic materials using a data-driven approach, where a neural network is trained to learn the non-linear relationship between boundary conditions and the resulting displacement field.  We also introduce a method to guarantee the validity of the solution. In total, we present three contributions: an optimized data set generation algorithm based on modal analysis, a physics-informed loss function, and a Hybrid Newton-Raphson algorithm. The method is applied to two benchmarks: a cantilever beam and a propeller. The results show that our network architecture trained with a limited amount of data can predict the displacement field in less than a millisecond. The predictions on various geometries, topologies, mesh resolutions, and boundary conditions are accurate to a few micrometers for non-linear deformations of several centimeters of amplitude.}
\maketitle

\section{Introduction}

Recently, with the increase in GPU computational power, machine learning started to revolutionise several fields, in particular computer vision, language processing, and image processing. Deep-learning is an area of machine learning that has demonstrated a strong ability at extracting high-level representations of the relation between a given input and its corresponding output as opposed to task specific algorithms. In other words, given enough inputs, such techniques can approximate the relation to the corresponding outputs without any prior knowledge.

Numerical methods such as the Finite Element Method \cite{zienkiewicz1977finite} (FEM) are vastly used in science to solve partial differential equations (PDE) on complex domains, for which analytical solutions are not possible. Computing the non-linear deformation of mechanical structures is one of the fields which deals with such equations and uses FEM to approximate the solution.
%The FEM is well suited for solving equations associated with non-linear materials and is the one that we consider in this article. 
The main benefits of the FEM are its accuracy and well-grounded mathematical foundations. %and ability to simulate a large range of materials on potentially complex domains.
However, when the problem complexity increases and accurate solutions are sought, the combination of high resolution meshes and non-linear constitutive laws usually leads to  computation times that are too high for certain applications.   

Our main objective here is to leverage the advantages of deep learning methods (in particular the ability to learn complex relations between inputs and outputs of a model) and the solid scientific foundations of the FEM to obtain very fast, yet accurate solutions of nonlinear elasticity problems. 
%The goal being of providing static deformations from various external forces with micro-scale error within the microseconds computation time. 
Since multiple strategies have been proposed over the last decades to reduce the computation time of FEM, the following sections describes some of these techniques and discusses about their limits compared to our expectations.

\paragraph{Domain decomposition}
As the name suggests, Domain Decomposition \cite{brenner2007mathematical} relies on a smart decomposition of the initial domain into multiple sub-domains. The associated (simpler) sub-problems are coupled via  disjoint or overlapping boundaries. Different algorithms have been proposed to solve such problems. Sub-structuring algorithms such as BNN \cite{mandel1993balancing} or FETI \cite{farhat2001feti} can only handle non-overlapping domain decomposition while others like Schwarz \cite{black1870ueber} or Lions \cite{lions1990schwarz} methods can work on both overlapping and non-overlapping sub-domains. Combined with parallel processing, it is possible to achieve significant speedups while maintaining good convergence properties \cite{haferssas:hal-02400982}. However, the optimal gain for Domain Decomposition methods is observed when dealing with very large problems (i.e. with millions of degrees of freedom) \cite{haferssas:hal-02400982}. On smaller resolution meshes, or when real-time computation is needed, the speedup obtained by this approach is more limited as the computation of boundary interactions becomes predominant \cite{haferssas2019simulation}. Since we want to favour fast or real-time computation of nonlinear elastic structures, domain decomposition techniques are not well suited.
%\textcolor{green}{The presented method uses a fully-connected neural network which has a quadratic growth in number of parameters. During the training phase, the ANN, the batch of data and the gradient of the network output has to fit on the GPU memory at the same time. With these memory space limitation we only consider objects up to 15000 degrees of freedom which makes this method unsuited for our application cases.}

\paragraph{Proper Orthogonal Decomposition}
Another approach to improve computation times is Model Order Reduction. Among this class of methods, Proper Orthogonal Decomposition (POD)  \cite{chatterjee2000introduction} has proved to perform well on a wide variety of problems ranging from fluid dynamics \cite{berkooz1993proper} \cite{mendez2019multi}, soft robotics \cite{goury2021real}, aeronautics \cite{macmanus2017complex}, optimal design \cite{oliveira2007reduced} and many others. This approach aims at reducing the number of Degrees of Freedom (DOFs) by analysing deformation modes and discarding the least significant ones. These modes are obtained via a "data set" of $P$ samples that are put together as a matrix $\vect{Q}$ of size $N \times P$ where $N$ is the number of DOFs. The eigenvalues and their associated eigenvectors are then computed from the self-adjoint matrix $\vect{Q}\vect{Q}^T$. When the displacement field is well behaved, the magnitude of the eigenvalues decreases quickly, showing that the object is mostly characterised by the first few modes of deformation (i.e. the  eigenvalues with the highest norm). Follows a step where the lowest eigenvalues (usually anything $10^{-8}$ times smaller than the magnitude of the first one) are removed. The simulation is computed in the reduced space thus speeding up the solving of the system. This method relies on a trade-off between accuracy and speed. There are multiple factors impacting the quality of the simulation and the gain such as how magnitudes of eigenvalues decrease and the magnitude of the cutoff value. Although it introduce errors in the simulation, the computed reduced space can be used to solve "similar" problems with slight changes in the boundary conditions or model parameters \cite{park1996use} \cite{maday2004reduced}. This allows to reuse the reduced model for a variety of scenarios, therefore reducing the overall computational cost of the method. Artificial neural networks have already proven to be resilient to geometry variations, Pfeiffer \textit{et al.} \cite{pfeiffer2019learning} trained a neural network on randomly generated meshes to predict displacement fields. We will demonstrate in this article that the presented approach is at least as fast as the POD while providing more flexibility to model parameters variations.

\paragraph{GPU-acceleration}
Many publications have addressed the problem of computational performance for FE simulations through GPU-accelerated approaches. While the use of a parallel implementation can speedup the computation of the local, element-wise, stiffness matrices, these methods essentially aim at solving the global linear system associated with the linearization of the problem.  

In the case of an iterative method, such as the conjugate gradient, most of the gain can be achieved by improving the sparse matrix-vector operations. Multiple methods have been explored to implement efficiently these operations on the GPU \cite{nvidia_gpu}. Mueller \textit{et al.} \cite{mueller2013matrixfree} or Martínez-Frutos \textit{et al.} \cite{MARTINEZFRUTOS20159} use the fact that CG iterations can be performed without explicitly assembling the whole matrix. This has the advantage of reducing the memory bandwidth while being fast and stable. As an example, Allard \textit{et al.} \cite{allard:inria-00589200} proposed a cache optimisation process for a co-rotated formulation of a linear elastic model. With this method, a mesh composed of 20k tetrahedra can be simulated in about 2 ms \cite{allard:inria-00589200}. This approach is however too specific to be applied to other materials such as hyperelastic ones. 

In the case of direct solvers, the assembly of the global matrix is required to compute the decomposition or factorisation of the system. %When combined with efficient solvers the assembly process can become the bottleneck. 
Dziekonski \textit{et al.} \cite{FEMONGPU} and Mueller \textit{et al.} \cite{mueller2018gpu} proposed to assemble the matrix directly on the GPU thus reducing memory transfer and speeding up the assembly. In this case also, the method requires a model specific algorithm and cannot provide a good combination of heterogeneous CPU/GPU simulations.
One could also optimise the construction of the matrices, using very efficient implementation such as the one provided by the Eigen\cite{eigenweb} library.
 
Overall, despite some limitations, GPU-based FEM algorithms provide fast and accurate results. Yet, as for the POD, GPU-based FEM falls under the category of task specific algorithms. It de facto encounters the same limitations when using different hyper-elastic laws to represent complex objects. 
%\SC{As an example, in the case of liver surgery simulation, achieving similar speedup  would require to define the hyper-elastic law of a patient specific organ if possible at all. The next step is to implement and optimise the equation on the GPU to finally run the simulation. } \COM{ $\longrightarrow$ on s'en doute un peu qu'il faut implementer et executer la simu non ?} \SC{Where from acquired data one could train a neural network as a black box to approximate the hyper-elastic law of the patient.} \COM{ $\longrightarrow$ à ré-écrire car il y a des référence à de la simulation de chirurgie dont on n'a jamais parlé avant} 

\paragraph{Physics Informed Neural Network}
Recently, Raissi et al. \cite{raissi2019physics} proposed a novel method for the approximation of the solution of PDEs using Artificial Neural Networks. Named Physics Informed Neural Networks (PINN), they leverage the recent developments in automatic differentiation \cite{baydin2018automatic} to differentiate neural networks with respect to their input coordinates and model parameters using an unsupervised learning process. This method relies on a novel formulation of the objective/loss function. In physics, problems are usually composed of three part: initial conditions, boundary conditions, and a governing equation. In order to satisfy the problem, a solution has to hold true the conditions while solving the governing equation. Raissi \textit{et al.} use this pattern to formulate the loss function where each part of the problem will be evaluated with the output of the network. The Mean Squared Error (MSE) computed for each part from the output is then summed up and, via an optimisation algorithm, the weights and biases of the network are updated. This approach has proved successful in several fields and application cases such as cardiac activation mapping \cite{10.3389/fphy.2020.00042}, modelling of fractures \cite{goswami2020transfer} or fluid dynamics \cite{w13040423}. PINN is a fast, efficient and elegant way of training a neural network to solve ordinary and partial differential equations. However, the approach constrains the inputs to be the differentiation variables. Furthermore it does not support variations of the boundary conditions since their formulation appears in the loss function. As a result, each time the problem is changed (i.e. different Neumann boundary conditions, different model parameters, etc.) the network has to be trained again. This limits significantly the benefit of the approach when compared, for instance, to GPU-based methods or POD techniques. Our method allows for the computation of various deformations from a single training process. The method uses external forces applied on the object as inputs of the network which are not the differentiation variables of the constitutive laws, thus, making the Raissi \textit{et al.}'s hypothesis not applicable in this case.
%In order to approximate a wide variety of displacement fields we specifically change the boundary conditions of the domain thus giving them as input of the network. 
%\SC{The next section presents the governing equation and shows that the input variables are not present in its definition making the Raissi \textit{et al.}'s hypothesis not applicable in this case.} \COM{ $\longrightarrow$ pas clair}

\paragraph{Data-driven learning process}
Data-driven methods are a class of machine learning techniques that can deal with a wide variety of physics-based problems ranging from radio-frequency or microwave modeling \cite{zhang2003artificial} to fluid mechanics \cite{kochkov2021machine} and solid mechanics \cite{mendizabal2020simulation}. 
%or even video enhancement \cite{chu2020tecoGAN}. 
Data are generated from real world acquisitions \cite{romaguera2020prediction} or can be computer-generated \cite{chentanez2020cloth}. 
%The data are then processed and considered noise free. 
Once the data are processed or generated, the inputs are fed to the network, and the outputs are compared to the ground truth via a loss function (usually the MSE). The error gradient is then computed and, finally, the weights and biases of the artificial neural network are updated using an optimization algorithm. 
Such an approach is helpful to approximate problems that do not have a mathematical model, such as image animation \cite{holynski2020animating} or problems that do not fit into the PINN framework.

Prior work has been done on the topic of simulating hyper-elastic materials using data-driven approaches. Cloth deformation is an important subject in the field of physics-based animation, and Wang \textit{et al.} \cite{wang2011data} used measured data to build a piece-wise bending and stretching model to compute the non-linear dynamic of cloth material. Xu \textit{et al.} \cite{xu2014sensitivity} following the idea of Kim \textit{et al.} \cite{kim2013near} proposed a technique of mix and match to generate meshes that fit the desired pose. More recently, Santestaben \textit{et al.} \cite{santesteban2019learning} used a neural network to generate non-linear garment wrinkles. For applications involving the deformation of elastic solids, Brunet \textit{et al.} \cite{brunet2019physics} proposed a method to compute hyper-elastic volume deformation of a liver in real-time from a set of Dirichlet boundary conditions. Finally, Meister \textit{et al.} \cite{meister2018towards} and Mendizabal \textit{et al.} \cite{mendizabal2020simulation} proposed neural networks able to predict the deformation of an elastic structure given variable external forces. The later approximate static deformations using a Convolutional Neural Network (CNN). This method requires to immerse the object within a topological grid in order to perform convolutions. Neumann boundary conditions are transferred to the grid through a mapping. The deformation is then computed on the grid by the network and applied on the object using the inverse mapping. These methods provide a fast estimations of the displacement field. Their principal drawbacks are related to the data generation process, and the limited accuracy of the solution. The amount of simulations required to train the network (several hundred simulations per boundary node) becomes time consuming when the grid resolution is increased, and so does the training time. The predictions accuracy is in the range $[10^{-4};10^{-3}]$ of the object scale, which is acceptable for some applications but not sufficient for others.
\newline
\newline
In this article we present an improvement of this latter class of methods. We first briefly describe the network architecture. In a second time we propose a novel technique to generate meaningful deformations based on modal analysis. To follow this, we present a physics-aware loss function that significantly improves the training and prediction accuracy. As our last contribution we propose a variation of the Newton-Raphson algorithm that uses the prediction of the ANN. Finally, we analyse and discuss the results of our method.

\section{Method}

The deep learning aided simulation can be split into two main techniques: the PINNS introduced by Raissi et al. \cite{raissi2019physics} in 2019 and the simulation-based approach. In this paper we focus on the latter, with the long-term objective to learn also from real data and to obtain as-generic-as-possible neural networks, able to predict elastic deformations for a wide range of cases.

\subsection{Neural network architecture}\label{networkArc}

In recent years, Deep Neural Networks (DNN) have led to many successful applications such as image recognition and natural language processing thanks to their exceptional expressibility. Starting from a simple feed forward fully-connected neural network, scientists have proposed more complex architectures such as Convolutionnal Neural Network (CNN), and combined multiple architectures to suit their problems and constraints. CNNs have shown some incredible results in their generalisation capabilities and high-level features extraction.

In a context similar to this paper, Mendizabal et al. \cite{mendizabal2020simulation} used a specific CNN architecture named Unet. This network takes as input a $3 \times n_x \times n_y \times n_z$ force tensor. It corresponds to the force interpolated on each node of the axis aligned bounding grid. It outputs a similarly sized tensor corresponding to the grid displacement at each nodes.

In this section we will demonstrate that even though CNN are a good and viable option in this context, simpler is better. With a simple Fully Connected Neural Network (FCNN) composed of two hidden layers, we manage to obtain an excellent prediction accuracy with a mean squared error between $10^{-7}$ and $10^{-10}$.

\subsubsection{Proposed Architecture}

\begin{figure}[ht]
    \begin{minipage}[c]{0.45\linewidth}
        A feed forward FCNN can be assumed to be the stack of the input layer, multiple hidden layers and the output layer. The connection between two adjacent layers can be expressed in the form of a tensor, as follows
        \begin{equation}
            \vect{z}_i = \alpha(\vect{W}_i\vect{z}_{i-1} + \vect{b}_{i}),\text{ for }1\leq i \leq n+1
        \end{equation}
        where $n$ is the total number of layers, $\alpha(.)$ denotes the activation function acting element wise, $\vect{z}_0$ and $\vect{z}_{n+1}$ denotes the input and output tensors respectively, $\vect{W}_i$ and $\vect{b}_i$ are the trainable weight matrices and biases in the $i^{th}$ layer. 
        
        Activation functions play an important role in the learning process of the neural networks. Their role is to apply a (nonlinear) function to the output of the previous layer and pass it to the next one.
        
    \end{minipage}\hfill
    \begin{minipage}[c]{0.50\linewidth}
        \centering
        \includegraphics[width=\linewidth]{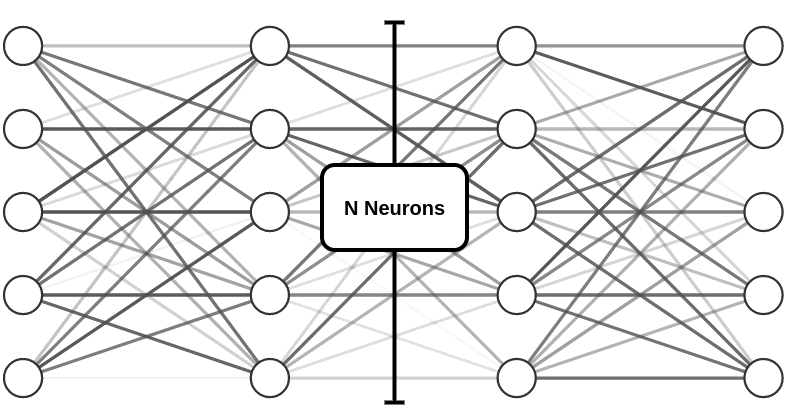}
        \caption{The proposed architecture is composed of 4 fully connected layers of size the number of dofs with a PReLU activation function.}
        \label{fig:FCNN}
    \end{minipage}
\end{figure}
\vspace{-2.0ex}
Only non-linear differential functions are considered since they allow for a useful information to be computed during the back propagation.
Here the activation function $\sigma(.)$ is a PReLU \cite{he2015delving}, which stands for Parametric Rectified Linear Unit. This function has multiple advantages such as: valuation and gradient computation speed, and neuron-wise adaptive activation.
\begin{equation*}
    \text{PReLU}(x) =
    \begin{dcases}
    x & x \in \mathbb{R}^+\\
    ax & x \in \mathbb{R}^-\ \  a \in \mathbb{R}^+\\
    \end{dcases}
\end{equation*}
\vspace{2.0ex}
The negative part has a learnable parameter $a$, allowing us to adaptively consider both positive and negative inputs.

Such architecture is easy to implement and fast, but has two drawbacks. The first is the number of parameters that need to be learnt by the network. This number is given by the polynomial $4N^2 + 7N$, where N is the number of degrees of freedom. The quadratic growth is a limitation for both GPU memory and  gradient computation. The second drawback of such an architecture is that FCNN are prone to over-fitting even though this phenomenon has not been observed in any of our scenarios.

\subsection{Mode-based data set generation}\label{datagen}

In simulation-based approaches, the generation of a synthetic data set  often represents an important part of the whole training process. The naive technique, which consists in applying random external forces at each node of the FE domain)  provides minimal control over the representativeness of the data set. Defining an optimal combination of external forces is difficult and depends on the model and has to be done mesh wise in order to obtain specific and interesting deformations.

Our method is based on modal analysis of the global stiffness matrix $\mathbb{K}^\prime$ arising from a finite element method. Thus, it can be applied with any constitutive model, mesh resolution or finite element type. While a modal analysis is used to produce the external forces, no reduction is performed as shown in equations \ref{Modal_base_formulation} and \ref{FMA2} thus preserving the physical properties of the model. It allows the computation of external forces that are responsible for the most significant deformations of a given mesh.

In order to explain both the method and the notations we start this section by a quick overview of the theory behind continuum mechanics. More detailed knowledge can be found, for instance, in \cite{wriggers2008nonlinear}. Finally, we demonstrate how to compute the external forces from the deformation modes. 

\subsubsection{Numerical simulation of hyper-elasticity problems}

For the analysis of nonlinear initial boundary value problem in continuum mechanics, a coupled system of partial differential equations has to be solved which consist of kinematics relations, local balance of momentum and the constitutive equations. We first present the hyper-elastic strong form and its alternative weak formulation Finally we proceed to present one of the approaches to solve it.

\begin{figure}[ht]
    \vspace{-2.0ex}
    \begin{minipage}{0.52\linewidth}
    The boundary-value problem reads as follows : find $\vect{d} : \bar{\Omega} \longrightarrow \mathbb{R}^{d}$
        \begin{equation}\label{eq:strong_form}
            \begin{aligned}[b]
            -\diverg \firstPiola(\disp) &= \vect{f} &\text{in }\Omega\\
        \disp &= \vect{g} &\text{on }\Gamma_{D} \subset \partial \Omega\\
        \firstPiola(\disp) \vect{n} &= \vect{h} &\text{on }\Gamma_{N} \subset \partial \Omega\\
        \firstPiola(\disp) \vect{n}  + \alpha \vect{u} &= \vect{0} &\text{on }\Gamma_{R} \subset \partial \Omega
            \end{aligned}
        \end{equation}
    \end{minipage}\hfill
    \vspace{-2.0ex}
    \begin{minipage}{0.46\linewidth}
       Where $\Omega \subset \mathbb{R}^{d} (d=2,3)$, $\partial \Omega$ the boundary of $\Omega$, $\firstPiola$ is the first Piola-Kirchhoff, $\vect{f}$ is the vector of the external forces, $\disp$ is the displacement vector, $\vect{g}$ a suitable Dirichlet's boundary conditions on $\Gamma_{D}$, $\vect{h}$ a Neumann's boundary conditions on $\Gamma_{N}$, $\vect{n}$ the normals on $\partial \Omega$, $\alpha$ the elastic coefficient associated with the Robin-Type boundary condition on $\Gamma_{R}$.
    \end{minipage}
\end{figure}

Let's introduce two functional spaces:
\begin{equation}
    \begin{aligned}[b]
        V &= \{\vect{w} \in \left[ H^{1}(\Omega)\right]^{d} : \vect{w}|_{\Gamma_D}=\vect{g}\}\\
        V_{D} &= \{\vect{w} \in \left[ H^{1}(\Omega)\right]^{d} : \vect{w}|_{\Gamma_D}=\vect{0}\}
    \end{aligned}
\end{equation}

The weak formulation of the boundary value problem \ref{eq:strong_form} reads:

Find $\disp \in V_{D}$ such that
\begin{equation}\label{eq:weak_form}
    \int_{\Omega} \firstPiola(\disp) : \grad{\vect{w}} \: d\Omega + \int_{\Gamma_{R}} \alpha\disp \cdot \vect{w} \: d\Gamma = \int_{\Omega} \vect{f} \cdot \vect{w} \: d\Omega + \int_{\Gamma_{N}} \vect{h} \cdot \vect{w} \: d\Gamma \ \ \ \forall \vect{w} \in V
\end{equation}

Let's introduce a Finite Element partition $\Omega_{h}$ of the domain $\Omega$ from which we construct a conforming FE space
\begin{equation}
    X_{h} = \left< x_1,\ldots,x_{N_{h}} \right> \subset \left[H^{1}(\Omega)\right]^{d}
\end{equation}
We also define $V_{h} = X_{h} \cap V$ and $V_{Dh} = X_{h} \cap V_{D}$. The finite element approximation of equation \ref{eq:weak_form} can be written as :

Find $\disp_{h} \in V_{Dh}$ such that:
\begin{align}\label{eq:FE_approx_weak_form}
    \int_{\Omega} \firstPiola(\disp_{h}) : \grad{\vect{w}} d\Omega + \int_{\Gamma_{R}} \alpha\disp_{h} \cdot \vect{w} d\Gamma &= \int_{\Omega} \vect{f} \cdot \vect{w} d\Omega + \int_{\Gamma_{N}} \vect{h} \cdot \vect{w} d\Gamma \ \ \ \forall \vect{w} \in V_{h}\\
    \vect{f}^{i}(\disp_{h}) &= \vect{f}^{e}\\
    \vect{R}(\disp_{h}) = \vect{f}^{i}(\disp_{h}) &- \vect{f}^{e} = \vect{0}\label{eq:residual}
\end{align}
Where $\vect{f}^{i}(\disp_{h})$ are the internal forces and $\vect{f}^{e}$ the external ones. $\vect{R}(\disp_{h})$ is called the residual vector.

From now on, we will only work with equation \ref{eq:FE_approx_weak_form} and in order to do not overload the notations we will denote $\disp \in \mathbb{R}^{N_h}$ the vector of coefficients in the expansion of $\disp_h$ with respect to the FE basis function $\varphi_i$.

Finding the roots of a function has been a hot topic for multiples centuries now. From these researches emerges the well studied Newton-Raphson process that has proven to be very effective at finding roots of non-linear equations. As of today, this algorithm and its variations are commonly used to solve FE problems and is the one we picked to solved such problems.
In the next subsection we will see how we use the tensor $\mathbb{K}(u)$ in order to obtain a given displacement, and its corresponding forces.

\subsubsection{Modal forces}\label{subsubsection:modal_force}

External force distribution on the object boundary (i.e. the choice of Neumann boundary conditions) has an direct impact on the resulting deformation, yet it is not always intuitive to know how to create a given displacement from a set of external forces. The opposite is also true, where two different distributions of forces can actually give a similar deformation. This is why it is difficult to generate an optimal training data set composed of a large variety of (non-redundant) deformations. For example, Mendizabal et al. \cite{mendizabal2020simulation} use a heuristics to generate the training data set, which requires to be tuned for each application in order to be computationally stable.

\begin{figure}[ht]
    \centering
    \includegraphics[width=0.80\linewidth]{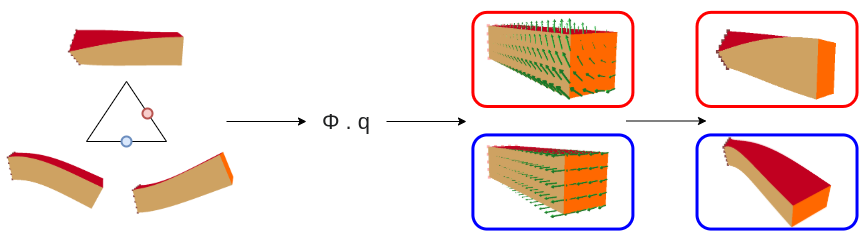}
    \caption{Example of a deformation basis and its resulting forces / linear deformations. $\Phi$ and $\vect{q}$ are the matrix and vector described in equation \ref{Modal_base_formulation}. The volume gain is a known artefact of linear elasticity laws when rotations are involved. }
    \label{fig:Linear_def}
\end{figure}

\begin{figure}[ht]
    \centering
    \includegraphics[width=0.9\linewidth]{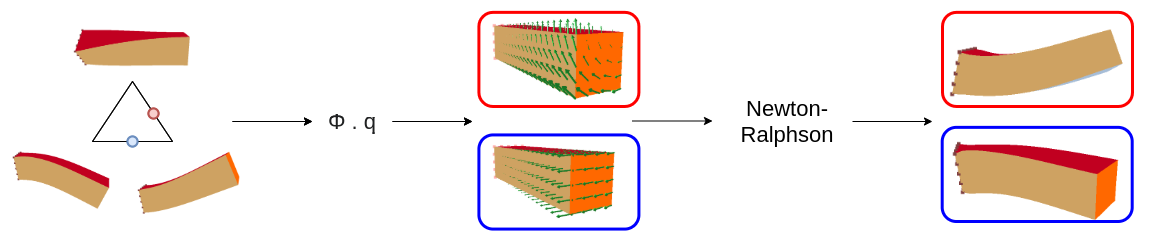}
    \caption{Example of a deformation basis and its resulting forces / non-linear deformations. $\Phi$ and $\vect{q}$ are the matrix and vector described in equation \ref{Modal_base_formulation}.}
    \label{fig:mod_force_hyperelas_def}
\end{figure}

One way to maximise the information that goes through the network would be to construct a deformation basis. This basis can then be interpolated within this space to obtain  an approximation of the constitutive law, as shown in Figure \ref{fig:Linear_def}.
\vspace{-1.0ex}
This might be sufficient for linear elasticity problems, but the hyper-elastic materials have a more complex relation between forces and displacements. For a non-linear elastic model, the stiffness matrix is itself a function of the DOF values (or their derivatives). We need an iterative approach, such as the Newton-Raphson method, to solve the nonlinear equations and compute the correct deformation caused by the external forces. This is described in figure \ref{fig:mod_force_hyperelas_def}.

In this subsection we explain how we generate this deformation basis and compute external forces from modes.
Modal analysis relies on the eigenvectors of a given matrix, in this case we consider the tangent stiffness matrix  $\mathbb{K}(\disp)$ from equation \ref{eq:tangent_stiff_matrix}. Each eigenvector corresponds to a deformation mode. As a result, any deformation of an elastic object can be represented as a combination of these vectors. Hyper-elasticity problems cannot be represented by a single matrix hence the system of equations is solved using a sequence of linearizations. Since this process is time consuming, we choose here to apply the modal analysis only the tangent stiffness matrix associated with the rest configuration of the elastic body.

The linearisation process of the Newton-Raphson (equation \ref{eq:Newton_raph_process} and \ref{eq:tangent_stiff_matrix}) leads to the equation 
\begin{equation}\label{FMA}
	\mathbb{K}(\disp^{k})\cdot \dispDelta^{k} = \vect{R}(\disp^k)
\end{equation}

For simplicity we only write $\mathbb{K}$ instead of $\mathbb{K}(\disp = \disp^{k})$. From its eigen decomposition $\mathbb{K} = \Phi \Lambda \Phi^{T}$ we can write
\begin{equation}\label{Modal_base_formulation}
\begin{split}
    \mathbb{K}\cdot \dispDelta^{0} &= \vect{R}(\disp^{0}) \iff\\
	\Phi \Lambda \Phi^{T}\cdot \dispDelta^{0} &= \vect{f}^{i}(\disp^{0}) - \vect{f}^{e} \quad \text{with} \quad \vect{R}(\disp^{k}) = \vect{f}^{i}(\disp^{k}) - \vect{f}^{e}\iff\\
	\Phi (\Lambda \Phi^{T}\cdot \dispDelta^{0}) &= \vect{f}^{i}(\disp^{0}) - \vect{f}^{e} \iff\\
	\Phi \cdot \act &= - \vect{f}^{e}  \quad \text{with} \quad \vect{f}^{i}(\disp^{0}) = \vect{0}
\end{split}
\end{equation}
where $\Phi$ is the matrix composed of the eigenvectors of $\mathbb{K}(\disp^0)$,  $\Lambda$ is the diagonal matrix composed of the corresponding eigenvalues of $\mathbb{K}$, and finally $\act = \Lambda \Phi^{T} \dispDelta^{0}$. Each coefficient of $\act$ reflect how much its associated deformation mode is present in the applied external force.  

The computation of the eigenvectors can be time consuming on very fine meshes, but is reasonable in the scenarios we are considering ( around 4 minutes and 20 seconds for 15,000 DOFs). Yet, as it occurs only once in the whole training phase hence, this computation time remains negligible when compared to the whole training process.

The selection of the first few modes (setting the first values of $\act$ to a non zero value) generates force vectors which correspond to the principal deformations of the object. 
We only consider a few modes to generate the external forces. The left-hand side of the equation is unchanged hence, the physical properties of the systems remain the sames. The equation simply becomes: 

\begin{equation}\label{FMA2}
	\mathbb{K}\cdot \dispDelta = \vect{f}^{i}(\disp) + \Phi\cdot\act
\end{equation}

The size of the training data set is then controlled by the sampling of those modes. From tests performed on various meshes we noticed that with only the first three to five modes and an overall sampling of 1,000 to 2,000 simulations we can create a learnable and generalisable data set leading to accurate predictions. To further improve the generalisation ability of the network, we apply the previously computed forces on random areas of random size over the mesh boundary. This gives us the ability to better represent local deformations.
With this approach the parameter tuning only consists on choosing the number of modes and the sampling of each mode. In the following sections we will consider that we have no prior knowledge on the use case scenario hence, we will use a uniform sampling of each mode. However, application knowledge can be used to select particular modes and specific sampling in order to have a more representative training data set. 

Depending on the material characteristics, one might need to apply very large or very small forces, to obtain the expected amplitude of deformation. Making a neural network robust to such extreme inputs and outputs is proven difficult, and we use a batch normalisation process on the input forces and output displacements to prevent this issue. 

\subsection{Loss function}

Deep learning uses the gradient of the error to learn from its previous mistakes. The error is computed via a function called loss function. There are various factors involved in choosing a specific loss function such as: ease of calculating the derivatives, the percentage of outliers in the data set, the type of machine learning algorithm, etc.
One of the most known and used function is the mean squared error (MSE).
\begin{equation}
    \text{MSE}(\disp) \defeq \frac{\sum_{i=0}^{N-1} (\disp_i-\vect{v}_i)^{2}}{N}
\end{equation}
Where $\disp$ is the value predicted by the ANN and $\vect{v}$ the ground truth. It has nice mathematical and computational properties and values at $0$ only when the output / prediction of the network exactly matches the expected value.
$\vect{v}$ is the ground truth divided by the length of the longest axis of the object-oriented bounding box in order to become scale independent.

\begin{figure}[ht]
    \begin{minipage}{0.48\linewidth}
        \centering
        \includegraphics[width=0.65\linewidth]{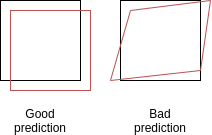}
        \caption{Example of two predictions (red) that have similar errors compared with the ground truth (black). The right-hand side prediction is 22\% more accurate than the left one according to the MSE, yet we would prefer the left one that better preserve some physical properties.}
        \label{fig:PReLU}
    \end{minipage}\hfill
    \begin{minipage}{0.48\linewidth}
        To improve the discriminative aspect of the loss function, and enforce its physics-based nature, we propose to rely on the internal forces of the elastic model. We know that at equilibrium the total amount of force in a system sums up to zero. Using equation \ref{FMA} we can then compute the \textit{residual forces} represented by the vector $\vect{R}(\disp)$ and we call residual its euclidean norm. This criterion and its relative counterpart describe with a good accuracy the state of the system.
        
        While we want to keep the geometric convergence from the MSE we also want to enhance it with some physical knowledge. From these statements two basic approaches emerge.
        \begin{align}
            \label{MSEPR}
            L_r ^+(\disp) &= \text{MSE}(\disp) + \frac{\norm{\vect{R}(\disp)}}{\norm{\Phi \act _{max}}}\\
            \label{MSETR}
            L_r ^*(\disp) &= \text{MSE}(\disp) \times \frac{\norm{\vect{R}(\disp)}}{\norm{\Phi \act _{max}}}
        \end{align}
        
    \end{minipage}
\end{figure}

However, the MSE metric being based solely on geometry, it does not enforce physically-correct configurations, as illustrated in figure \ref{fig:PReLU}. This can introduce important errors in the prediction when complex deformations take place. Since the MSE alone is not a good enough discriminant for such problems, another metric needs to be proposed.

Where $\act _{max}$ is given by $\underset{i}{\max} \norm{\act_i}$. We recall that $\Phi$ is the matrix composed of the eigenvectors of $\mathbb{K}$.
Although numerically correct, the formulation of the $L_r ^+$ loss poses some issues when it comes to its implementation. In our approach, and others dealing with similar problems, the simulations are computed using an external software library. In order to evaluate the residual forces, outputs of the ANN have to be cast in the corresponding framework type. During that process, we loose the derivation graph previously computed using autograd \cite{autograd}. Due to this lack of derivation graph, during the back propagation, the gradient of the normalised residual cannot be computed.

The formulation of the $L_r ^*$ loss fixes previously mentioned problems. During the computation of the back propagation, and using the linearity of the gradient, the normalised residual will be multiplied to the gradient of the MSE, thus, its value will propagate through the network. The normalised residual acts as a meaningful non-constant multiplicative coefficient which positively affect the learning process. Furthermore, the normalised residual is robust to scale-independent optimisation algorithms such as ADAM \cite{kingma2014method} since its value differs at each batch of data. 

In general, one can enhance the MSE loss function by any combination of the loss and the residual as long as the dimension analysis allows it. We want to bring the attention of the reader on the fact that the behaviour of the differential is also important and has to be taken into account in the conception process. In our case, and for the remainder of this method and associated results, we will use the $L_r ^*$ loss function.

\subsection{Hybrid Newton-Raphson algorithm}\label{HNRalg}

Since we are interested in the simulation of hyper elastic materials, we need a method that can handle nonlinear systems of equations. The Newton-Raphson algorithm (NRA) is a method often used technique for iteratively solving non-linear problems. %The main idea of the NRA is to construct a sequence $(\disp^n)$ using the gradient of a function $\vect{F}$ such that $\vect{F}(\disp^{\infty}) = 0$. The sequence is given by: 
%\begin{equation}
%\disp^0 = 0,\quad \quad \disp^{k+1}=\disp^{k}-\mathcal{J}^{-1}_{\mathcal{F}(\disp^{k})}\mathcal{F}(\disp^{k})=\disp^{k}+\dispDelta^{k}
%\end{equation}
One can present the process as follow:
\newline
We define the displacement $\disp$ as the limit of a sequence $(\disp^k)_{k \in \mathbb{N}}$ such that:
\begin{equation}
    \mathcal{F}(\disp) = 0 \quad \text{with} \quad \disp^0 = 0
\end{equation}
Then the Newton-Raphson process is given by
\begin{equation}\label{eq:Newton_raph_process}
(\disp^{k}-\disp^{k+1})=\mathcal{J}^{-1}_{\mathcal{F}(\disp^{k})}\mathcal{F}(\disp^{k})
\end{equation}
which is the same as 
\begin{equation}
\disp^{k+1}=\disp^{k}-\mathcal{J}^{-1}_{\mathcal{F}(\disp^{k})}\mathcal{F}(\disp^{k})=\disp^{k}+\dispDelta^{k}
\end{equation}
Using equation \ref{eq:residual} we have :
\begin{align}
    \mathcal{F}(u^{k}) &= \vect{R}(\disp^{k})\\
    \mathcal{J}_{\mathcal{F}(\disp^{k})} &= \frac{\partial \vect{f}^{i}(\disp^{k})}{\partial \disp} = \mathbb{K}(u^{k})\label{eq:tangent_stiff_matrix}
\end{align}
In the case of non-linear elasticity problems the jacobian of such function $\mathcal{F}$ is called the tangent stiffness matrix noted, $\mathbb{K}$.
Under the right conditions, the method converges and its convergence rate is quadratic. An example of good condition could be that the initial guess is close to the actual solution and there is no stationary point within the $(\disp^k)_{k \in \mathbb{N}}$ sequence. In order to set optimal convergence conditions for the Newton-Raphson algorithm, we use predictions from our neural network. For now, let us consider the classical method. It can be rewritten as: 
%The Newton-Raphson sequence can be rewritten as 
$\mathcal{J}_{\mathcal{F}(\disp^{k})}\cdot(\disp^{k+1}-\disp^{k})=-\mathcal{F}(\disp^{k})$. 
Therefore, the Newton-Raphson algorithm reads: 
\begin{figure}[ht]
    \vspace{-2.0ex}
    \begin{minipage}{0.48\linewidth}
        \begin{algorithm2e}[H]
            \SetAlgoLined
            \DontPrintSemicolon
            \KwData{$k=0$,$\ \disp^0 = 0$,$\ \dispDelta = 0$}
            \Repeat{$\norm{\vect{R}^k} < \epsilon$ or $\norm{\dispDelta} < \eta$}
            {
                Compute $\mathbb{K}(\disp^k)$\;
                Compute $\vect{R}(\disp^{k})$\;
                Solve $\mathbb{K}(\disp^k) \cdot \dispDelta =\vect{R}(\disp^{k})$   for $\dispDelta$\;
                $\disp^{k+1} = \disp^{k} + \dispDelta$\;
                $k = k+1$
            }
            \caption{Newton-Raphson algorithm}
        \end{algorithm2e}
    \end{minipage}\hfill
    \vspace{-2.0ex}
    \begin{minipage}{0.48\linewidth}
        It exists multiple options in order to speedup the Newton-Raphson algorithm. One can choose to improve the computation speed of $line\ 3$ using parallel computation as an example. Usually the bottle neck of the algorithm is at $line\ 4$ where most of the time is spent inverting the matrix $\mathbb{K}(\disp^k)$ which is composed of the number of dofs squared coefficients. Multiple approaches have been proposed to improve such computation. The Quasi-Newton-Method \cite{broyden1967quasi} create a matrix $\mathcal{B}$ which is an approximation of $\mathcal{J}^{-1}_{\vect{F}(\disp^{k})}$, recently, Duff \textit{et al.} \cite{duff2020new} proposed a new formulation of the $LDL^T$ solver using an \textit{a posteriori} threshold pivoting.
    \end{minipage}
\end{figure}

Finally one can also reduce the number of iterations needed to satisfy the condition on $line \ 2$ using a good initial guess $\disp^0=\disp_\vect{p}$. In this work we choose this approach, and use the prediction of the network to set an optimal starting point. This reduces the number of iterations of the algorithm most of the time, while also guaranteeing that we obtain a correct solution to our problem even if the prediction is not accurate or even incorrect.

The Hybrid Newton-Raphson-Algorithm reads as follows: 
\vspace{-2.0ex}
\begin{figure}[ht]
    \vspace{-2.0ex}
    \begin{minipage}{0.48\linewidth}
        \begin{algorithm2e}[H]\label{alg:HNR}
            \caption{Hybrid Newton-Raphson algorithm}
            \label{HNRA}
            \SetAlgoLined
            \DontPrintSemicolon
            \KwData{$k=0$,$\ \disp^0 = 0$,$\ \dispDelta = 0$}
            Compute $\vect{R(\disp^{0})}$\;
            $\disp_{p} = Prediction(\,R(\disp^{0})\,)$\;
            Compute $\vect{R(\disp_{p})}$\;
            \If{$\norm{\vect{R(\disp_{p})}} < \epsilon$}
            {
                exit\;
            }
            \Repeat{$\norm{\vect{R}(\disp^{k})} < \epsilon$ or $\norm{\dispDelta} < \eta$}
            {
                Compute $\mathbb{K}(\disp^k)$\;
                Compute $\vect{R}(\disp^{k})$\;
                Solve $\mathbb{K}(\disp^k) \cdot \dispDelta =\vect{R}(\disp^{k)}$   for $\dispDelta$\;
                $\disp^{k+1} = \disp^{k} + \dispDelta$\;
                \If{$k = 0$ and $\norm{\vect{R}(\disp^{1})} > \norm{\vect{R}(\disp_{p})}$}
                {
                    $\disp^1 = \disp_{\vect{p}}$
                }
                $k = k+1$
            }
        \end{algorithm2e}
    \end{minipage}\hfill
    \vspace{-2.0ex}
    \begin{minipage}{0.48\linewidth}
        The logic of the algorithm remains unchanged. There are multiple scenarios to consider, first the trivial one where $\disp_{\vect{p}} = \disp^{0}$, algorithm \ref{HNRA} introduces a slight computational overhead at $line\ 3$ but produces the same answer. The second scenario is the one where $\norm{\vect{R}(\disp^{1})} <= \norm{\vect{R}(\disp_{p})}$ at $line\ 12$. In this case the prediction does not provide any gain to the simulation. This could be due to $\vect{R(\disp^{0})}$ being too different from the training data. The artificial neural network cannot generalise enough to produce a good answer. This can also happen if the force is too small and produces a displacement field in the order of magnitude of the noise generated by the network. The third and last scenario is when the condition at either $line\ 4$ or $line\ 12$ is satisfied. In the best case when it stops at $line\ 4$ we can compute the displacement field in a couple of milliseconds. In the other case, the prediction usually reduces the number of iterations needed  to satisfy the condition on $line\ 7$, thus speeding up the algorithm when compared to its classical version.
        
        In the next section we will discuss the results of the method and more specifically the Hybrid Newton-Raphson algorithm in subsection \ref{HNRA_results}.
    \end{minipage}
\end{figure}
\vspace{-2.0ex}

\section{Results}\label{result}

In this section, we start by presenting the chosen metrics to study our results. In order to assess the different assumptions made in this article, the method will be applied over four different experiments.
The first experimentation aims at showing the variety of use cases in which our method works. To do so, we discuss the prediction errors of two use cases that differ by the meshes, the hyper-elasticity law, the young modulus and the scale.
The second experimentation deals with the impact of the residual in the loss. Both ANN will have the same starting set of weights, data set, and training process. The only difference will appear in the formulation of the loss function where one uses a classic MSE while the other uses $L_r^*$.
The third experimentation explores the generalisation abilities of the ANN. We consider a similar scenario as previously shown, but here, simulations and predictions are computed with random Young's modulus.
The fourth experimentation investigates the weaknesses and strengths of the Hybrid Newton-Raphson algorithm presented in the previous section. We will consider extreme cases with important and dim forces and also the average use cases.
\subsection{Metrics}
We use three different metrics to assert performances of the artificial neural networks. Those values are computed from 100 independent simulations.

\subsubsection{Maximum relative L2}
The first metric used is the maximum relative L2 error noted $e_{max}$. It gives the distance of the 3D node that is the furthest away from the ground truth over the 100 simulations.
\begin{equation}
    e_{max} = \max\limits_{s \in [1,S]}\big(\max\limits_{i \in [1, M]}\norm{\disp_i^s-\vect{v}_i^s}\big)
\end{equation}
Where $M$ is the number of 3D nodes, $S$ the number of independent simulations, $\disp$ the output of the network and finally $\vect{v}$ the solution computed by the solver. While this metric is a good indicator of how wrong the worst part of the object is deformed, it does not take into account more global phenomenons. Thus, it cannot be the sole metric in our analysis.

\subsubsection{Mean relative L2}
The second metric is the relative mean L2 error noted $e_{mean}$. It gives us the average per dofs relative error.
\begin{equation}
    e_{mean} = \frac{1}{N \times S} \sum_{s=0}^{S-1} \norm{\disp^s-\vect{v}^s}
\end{equation}
Where $N$ is the number of dofs, $S$ the number of independent simulations, $\disp$ the output of the network and finally $\vect{v}$ the solution computed by the solver. This metric is used in order to compute global information about the deformation error. This, combined with the previously detailed loss allows for a better understanding of the type of inaccuracy that the neural network provides. Yet, for inaccurately predicted small deformations and an accurately predicted big deformation, the previously mentioned error metrics will have low values. In order to properly interpret these low values and their relevance, a last metric is needed.

\subsubsection{Signal-to-noise ratio}
The third and last metric used is the signal-to-noise ratio noted SNR. It quantifies the power of noise/inaccuracy in a given signal.
\begin{equation}
    \text{SNR}^{dB}_{min} = \min\limits_{s \in [1,S]} 10 \times \log_{10}\big(\frac{\norm{\disp_s}}{\norm{\disp_s - \vect{v}_s}}\big)
\end{equation}
Where $S$ is the number of independent simulations, $\disp$ the output of the network and finally $\vect{v}$ the solution computed by the solver.
This metric will allow us to differentiate between the two previously mentioned problematic cases. Given a certain accuracy of the prediction, small displacements will provide small SNR while bigger displacements will provide bigger SNR.

\subsection{Experiments}
The following experiments will be realised over 100 randomly generated force samples. 
Given a mesh and a (hyperelastic) material law, we start a training using the previously defined process. Once the network is trained, each prediction is then compared to its corresponding FEM simulation, which serves as a ground truth. In order to work with comparable values, all the predictions $\disp$ and ground-truth $\vect{v}$ are scaled using the length of the longest side of the object-aligned bounding box noted $L$.

\subsubsection{General results}\label{gen_res}
This section discusses about the ability of the network to approximate deformations on various geometries with different elastic laws. To highlight the ability for our approach to learn on very different shapes and material properties, we selected a propeller mesh with a Neo-Hookean hyper-elasticity law and a cantilever beam with a Saint-Venant-Kirrchhof hyper-elasticity law. Both have about 12,000 degrees of freedom. Their material parameters are, on the other hand, very different with a low stiffness for the beam model, and very high stiffness for the propeller, in order to assess the validity of the training process and the predictions. Dimensions of the two structures were also chosen to be quite different for the same reason. 
The beam which is extremely soft provides displacements in the order of tens of meters from small external forces. Such extremes scenarios are usually the weak point of artificial neural networks.
In addition, given its geometry, it tends to deform globally even from local forces. An accurate prediction requires that the network can transfer the necessary deformation information to all the nodes.
On the opposite, for the propeller, high forces are needed to provide relatively small displacements, in the order of the centimetre. Yet, its geometry is such that with local forces it displays blade-wise deformations without any displacement on the neighbouring blades. This is a another important test for the neural network.
The different parameters of the 2 test cases are summarised in table \ref{tab:gen_results}. Both training lasted about 12 hours (including data set generation) on an NVidia TITAN RTX, with 4095 generated samples each.

\begin{table}[h]
    \centering
    \begin{tabular}{|c|c|c|c|c|c|c|c|c|}
    \hline
         Name & \#DOFs & Hyper-elastic law & L [m] & E [Pa] & Time [ms]& $e_{mean}$ & $e_{max}$ & $\text{SNR}^{dB}_{min}$\\
         \hline\hline
         Beam & 12,000 & Saint-Venant-Kirchhoff & 100 & $4.5 \times 10^{3}$ & 0.4 & $8.0 \times 10^{-6}$ & 0.03 & 8.4 \\
         Propeller & 12,075 & Neo-Hookean & 1.0 & $2.03 \times 10^{11}$ & 0.4 & $2.7 \times 10^{-6}$ & 0.01 & 18.9\\
         \hline
    \end{tabular}
    \captionof{table}{Results of a comparison between FEM simulations and ANN predictions over 100 randomly distributed forces with random amplitudes.}\label{tab:gen_results}
\end{table}
Both models having complete different sets of simulation parameters provide similar mean and max error over 100 simulations. This, although being demonstrated only on two models in this article, can argue the point that the network and its framework provide accurate global and local deformations of a mesh while handling a wide range of simulation parameters. The displacement field is computed in a steady $0.4 ms$ which is up to three orders of magnitude faster than the reference FEM simulation. To compare, the simulation of a deformation of the propeller takes around $500 ms$ to compute. One hypothesis to explain the big difference in SNR relies on the amount of near null deformations seen by both models. The beam has few samples where the displacement of the whole body is almost null hence is less trained at generating value close to 0. Where with the propeller most of the samples require a null or almost null displacement field for the vast majority of the points. A proper in-depth analysis on multiple models is required in order to conclude on this hypothesis. Works on this perspective are currently being held.
\begin{figure}[h]
\begin{subfigure}{.26\textwidth}
  \centering
  \includegraphics[width=\linewidth]{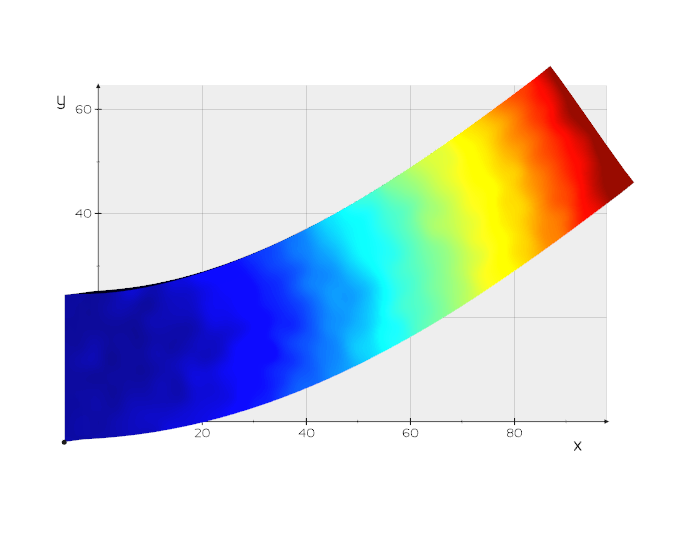}
  \caption{}
  \label{beam_max_def_up}
\end{subfigure}
\hfill
\begin{subfigure}{.26\textwidth}
  \centering
  \includegraphics[width=\linewidth]{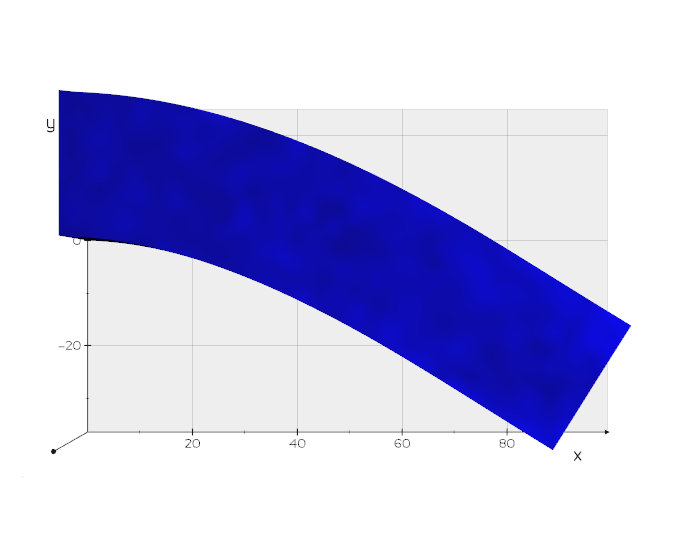}
  \caption{}
  \label{Beam_down_error}
\end{subfigure}
\hfill
\begin{subfigure}{.20\textwidth}
  \centering
  \includegraphics[width=\linewidth]{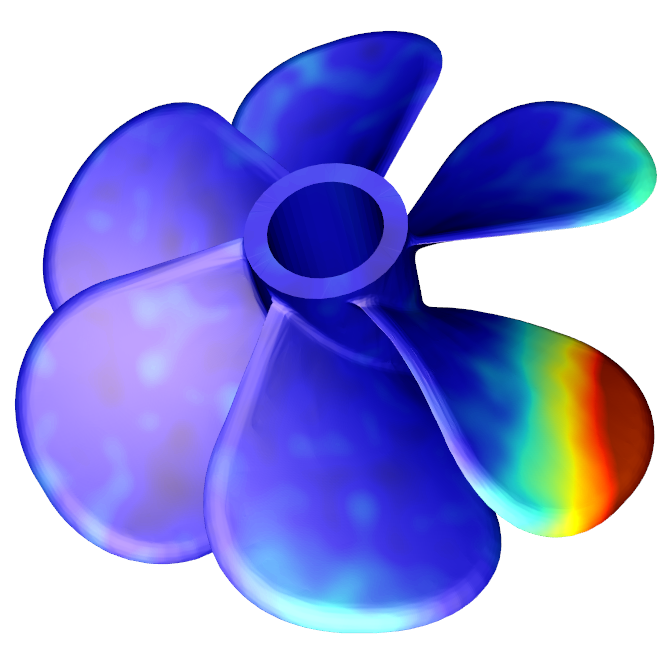}
  \caption{}
  \label{Propeller_big_error}
\end{subfigure}
\hfill
\begin{subfigure}{.20\textwidth}
  \centering
  \includegraphics[width=\linewidth]{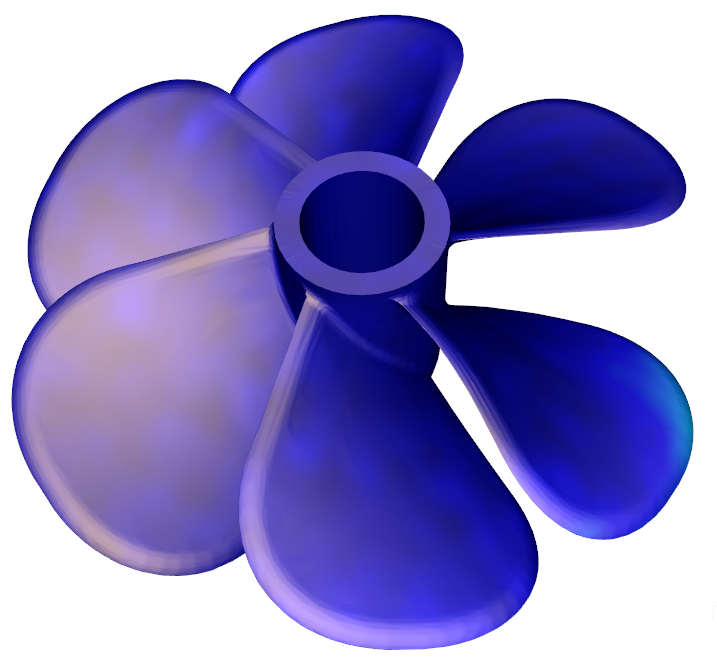}
  \caption{}
  \label{Propeller_small_error}
\end{subfigure}
    \caption{Examples of large nonlinear elastic deformations predicted by our neural network. The colours represent the node-wise euclidean distance to the solution of the Newton-Raphson algorithm. For both beams the colour gradient goes from $3\times10^{-4}$m (blue) to $3\times10^{-2}$m (red), for both propellers the colour gradient goes from $3\times10^{-5}$m (blue) to $2\times10^{-3}$m (red).}
    \label{fig:beam_and_propeller}
\end{figure}

\begin{figure}
\begin{minipage}{0.48\linewidth}
    \centering
    \begin{tabular}{|c|c|c|c|}
    \hline
         Name & $e_{mean}$ & $e_{max}$ & $\text{SNR}^{dB}$ \\
         \hline\hline
         Beam (a) & $5.4\times10^{-5}$ & $0.037$ & 22.2  \\
         Beam (b) & $6.6\times10^{-6}$ & $0.0050$ & 40.2  \\
         Propeller (c) & $2.82\times10^{-6}$ & $0.0036$ & 33.9  \\
         Propeller (d) & $8.2\times10^{-7}$ & $0.0008$ & 42.3  \\
         \hline
    \end{tabular}
    \captionof{table}{Error values and SNR of the deformations shown at figure \ref{fig:beam_and_propeller}. The displayed deformations are highly non-linear, yet the error values and SNR remain in the range of values displayed at table \ref{tab:gen_results}. }
\end{minipage}
\hfill
\begin{minipage}{0.48\linewidth}
    Beams (a) and (b) in figure \ref{fig:beam_and_propeller} undergo roughly the same amount of deformation (with a deflection at the tip of $\approx40$m) yet lead to very different error values and patterns. Beam (a) has an maximum error of 37mm near the free end of the beam reducing gradually to 54 µm near the fixed end. Beam (b) on the other hand provides an homogeneous error with a maximum error of 5.0 mm and an average error of 6.6 µm. The same behaviour can be observed with the propeller model. Propeller (c) while having a similar deformation than (d), has and average prediction error of 2.82 µm  with a maximum value at 3.6 mm, about four time as much as in the second scenario.
\end{minipage}
\end{figure}

\subsubsection{Physics aware loss function}\label{phy_aware}
Loss functions play an important role in the learning process of an artificial neural network. It exists multiple formulations for such functions and one can choose the one that suits better his data set and problem. When dealing with physics, loss functions that only rely on geometric difference might not be sufficient to properly train an artificial neural network. The formulation of the $L_r^*$ at equation \ref{MSETR} provides physical knowledge about the validity of the deformation by introducing the norm of the residual forces generated by the ANN prediction. The experiment will compare the accuracy of two ANN that only differ by the loss function used during the training (same initial weight, data set, normalisation process, etc...).
The ANN is trained on a $100 \times 25 \times 25$ meters beam composed of 12,000 DOFs with a Neo-Hookean hyper-elasticity law and a Young's modulus of 4,500 Pa. Both training took about 12 hours of computation (including data set generation)on a NVidia TITAN RTX, with 4095 generated samples each.

\begin{table}[h]
    \centering
    \begin{tabular}{|c|c|c|c|}
    \hline
         Loss function & $e_{mean}$ & $e_{max}$ & $\text{SNR}^{dB}_{min}$ \\
         \hline\hline
         $L^{*}_ {r}$ & $7.7 \times 10^{-6}$ & 0.03 & 2.7\\
         MSE & $5.2 \times 10^{-5}$ & 0.05 & -5.1 \\
         \hline
    \end{tabular}
    \caption{Results of a comparison between FEM simulations and the ANN predictions over 100 randomly distributed forces with random amplitudes. This table describes the accuracy of two ANN that only differ by the loss function used during the training.}
    \label{tab:LRSTAR_MSE_TABLE}
\end{table}
\vspace{-4ex}
\begin{figure}[ht]
\vspace{-2ex}
    \begin{subfigure}{.24\textwidth}
      \centering
      \includegraphics[width=\linewidth]{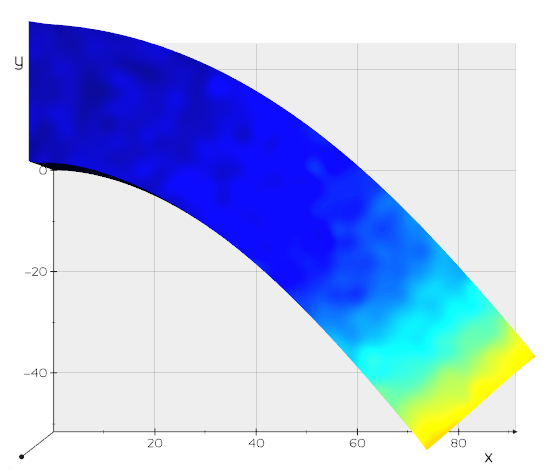}
      \caption{}
      \label{LRSTAR_beam_up}
    \end{subfigure}
%\vspace{-2ex}
    \begin{subfigure}{.24\textwidth}
      \centering
      \includegraphics[width=\linewidth]{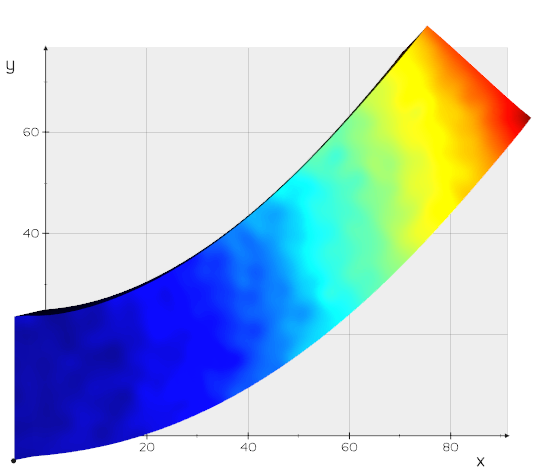}
      \caption{}
      \label{LRSTAR_beam_down}
    \end{subfigure}
%\vspace{-2ex}
    \begin{subfigure}{.24\textwidth}
      \centering
      \includegraphics[width=\linewidth]{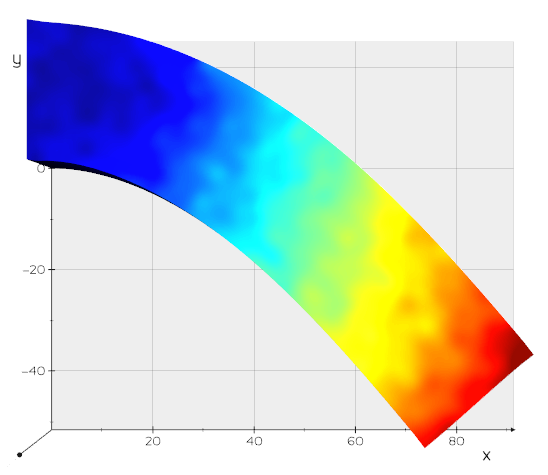}
      \caption{}
      \label{MSE_beam_up}
    \end{subfigure}
%\vspace{-2ex}
    \begin{subfigure}{.24\textwidth}
      \centering
      \includegraphics[width=\linewidth]{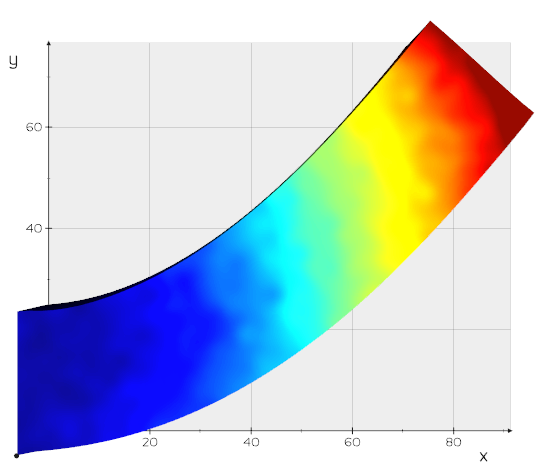}
      \caption{}
      \label{MSE_beam_down}
    \end{subfigure}
\caption{Nonlinear elastic deformations of a beam model predicted by our artificial neural network. The colours represent the node-wise euclidean distance to the solution of the Newton-Raphson algorithm. For all four configurations, the colour scale goes from $3\times10^{-4}$m (blue) to $3\times10^{-2}$m (red).}\label{fig:MSE_Lr_star}
\end{figure}

While the prediction of the neural network is very good in all cases, we can notice a difference due to the choice of loss function. Both networks have been trained from the same set of starting weights, data set, shuffle seed, normalisation process, etc., the only difference between the two training procedures being the loss function. The ANN trained using the $L^{*}_ {r}$ loss function provides results similar to the ones obtained in section \ref{gen_res} which display a certain stability in the training capacities of the method. The overall performances of the network trained using the proposed method are up by five to ten folds. The average error is the one that provides the best gain with a factor of 10, going down from $5.2 \times 10^{-5}$ to $7.7 \times 10^{-6}$. The maximum error sees an improvement by a factor of five while the noise is reduced by 6. The change in loss function has a strong impact over the learning of an ANN improving both its accuracy and generalisation abilities.

\begin{figure}[ht]
    \begin{minipage}{0.44\linewidth}
        \centering
        \begin{tabular}{|c|c|c|c|}
        \hline
             Name & $e_{mean}$ & $e_{max}$ & $\text{SNR}^{dB}$ \\
             \hline\hline
             $L^{*}_{r}$ (a) & $6.1\times10^{-6}$ & $0.027$ & 38.21  \\
             $L^{*}_{r}$ (b) & $1.5\times10^{-5}$ & $0.034$ & 34.7  \\
             MSE (c) & $5.3\times10^{-5}$ & $0.034$ & 25.42  \\
             MSE (d) & $5.5\times10^{-5}$ & $0.037$ & 24.9  \\
             \hline
        \end{tabular}
        \captionof{table}{Error values and SNR of the deformations shown in figure \ref{fig:MSE_Lr_star}. The displayed deformations are highly non-linear, yet the error values and SNR remain in the range of values displayed at table \ref{tab:LRSTAR_MSE_TABLE}.}
    \end{minipage}
\hfill
    \begin{minipage}{0.54\linewidth}
        The deformations displayed at figure \ref{fig:MSE_Lr_star}, the Beams (a) and (b) are both predicted using the neural network trained with the $L_{r}^{*}$ loss function while Beams (c) and (d) are predicted using the network trained with the MSE loss function. All predictions display similar pattern where the error increases toward the tip of the beam. The variations appear in the error metric where predictions (a) and (b) perform better overall than predictions (c) and (d). Forces applied on the beams are very important compared to what has been seen during the training process and one can notice that the prediction (c) and (d) remain around the mean values displayed in the table \ref{tab:LRSTAR_MSE_TABLE} while predictions (a) and (b) display errors up to two times bigger than average.
    \end{minipage}
\end{figure}

\subsubsection{Young's modulus generalisation}\label{young_generalisation}
The simulation has multiple parameters that can greatly influence the physics of an object. One of which is the Young's modulus. This positive coefficient represents the stiffness of the objects with values such as $1.96 \times 10^{11} Pa$ (iron) or $6.0 \times 10^{6} Pa$ (collagen). Given a force and a model, an increase in the Young's modulus produces smaller displacements and vice versa. Methods such as Model Order Reduction can solve problems with "slights" variations in simulation parameters \cite{park1996use} \cite{maday2004reduced}. The goal of this section is to show that the network can work with a wide variety of Young's modulus thus bypassing MOR limitations regarding this parameter. The ANN is trained on a $1 \times 0.25 \times 0.25$ meters beam composed of 12000 dofs with a Neo-Hookean hyper-elasticity law. The training lasted for 47h (including data set generation) on a NVidia TITAN RTX, with 10650 generated samples each.
\begin{table}[h]
    \centering
    \begin{tabular}{|c|c|c|c|}
    \hline
         Young's Modulus & $e_{mean}$ & $e_{max}$ & $\text{SNR}^{dB}_{min}$\\
         \hline\hline
         [$1.0 \times 10^{9}$, $2.03 \times 10^{11}$] & $4.6 \times 10^{-5}$ & 0.09 & -2.6\\
         \hline
    \end{tabular}
    \caption{Results of a comparison between FEM simulations and ANN predictions over 100 randomly distributed forces with random amplitudes and random Young's modulus within the given range.}\label{tab:Young_modulus_gen}
\end{table}
During the training, each computed sample is generated with a random Young's modulus between $1.0 \times 10^{9}$ and $2.03 \times 10^{11}$. The input tensor is then modified by adding a coefficient corresponding to the normalised value of the Young's modulus. The Young's modulus generalisation displays some losses compared to the presented results in section \ref{gen_res} but remain comparable to the MSE trained network presented in section \ref{phy_aware}. With an average error of $4.6 \times 10^{-5}$ it performs on average better than the MSE, but displays bigger max error with a $0.09$ against $0.05$ for the later.

When the range of Young's modulus becomes too important, the network cannot learn to generate deformations and tend to only create noise. With a range of [$1.0 \times 10^{6}$, $2.03 \times 10^{11}$] the network barely reaches the $1.0 \times 10^{-4}$ at the end of the training thus showing that it requires much more samples to converge to more accurate predictions.
\begin{figure}
\begin{subfigure}{.24\textwidth}
  \centering
  \includegraphics[width=\linewidth]{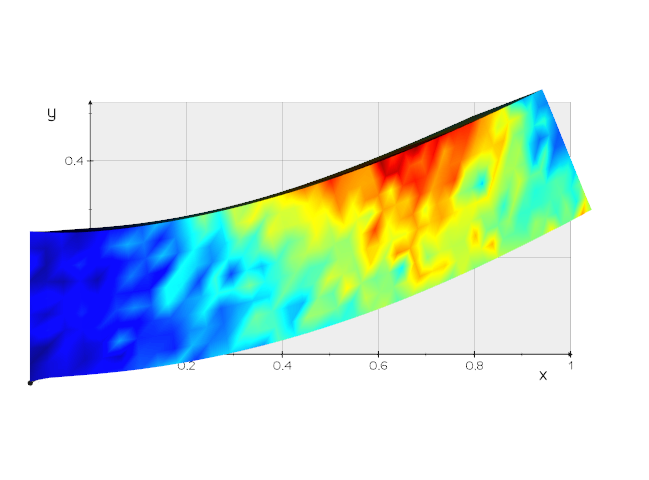}
  \caption{}
  \label{beam_1_multiple_E}
\end{subfigure}
\begin{subfigure}{.24\textwidth}
  \centering
  \includegraphics[width=\linewidth]{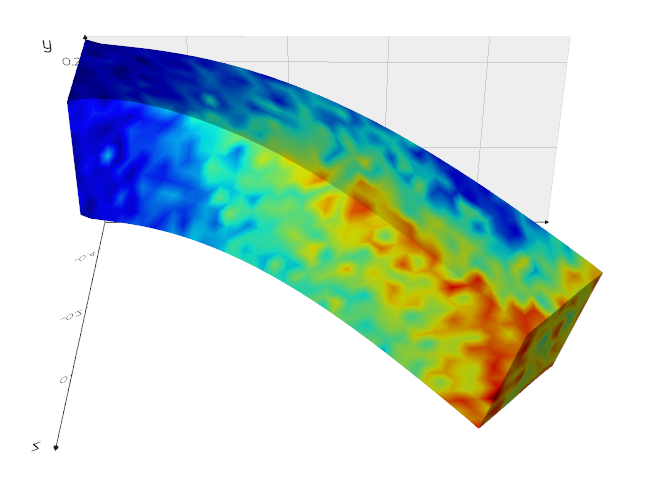}
  \caption{}
  \label{beam_2_multiple_E}
\end{subfigure}
\begin{subfigure}{.24\textwidth}
  \centering
  \includegraphics[width=\linewidth]{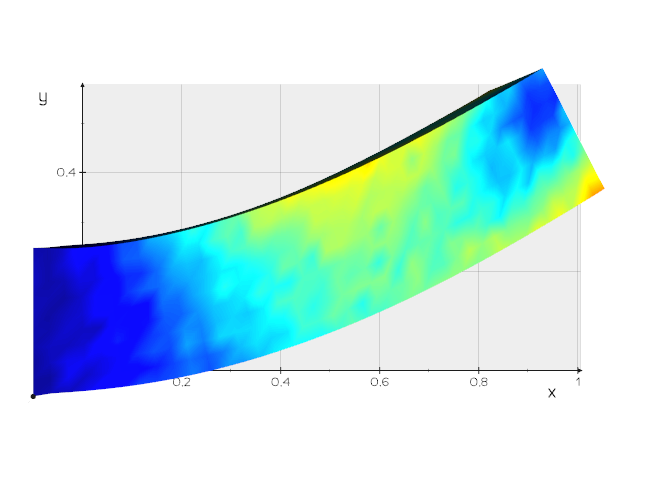}
  \caption{}
  \label{beam_3_multiple_E}
\end{subfigure}
\begin{subfigure}{.24\textwidth}
  \centering
  \includegraphics[width=\linewidth]{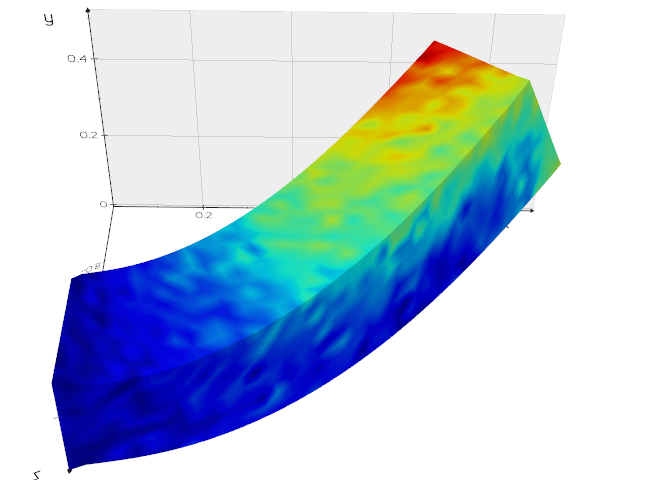}
  \caption{}
  \label{beam_4_multiple_E}
\end{subfigure}
    \caption{Nonlinear elastic deformations of a beam model predicted by our artificial neural network. The colours represent the node-wise euclidean distance to the solution of the Newton-Raphson algorithm. For all four configurations, the colour scale goes from $3\times10^{-4}$m (blue) to  $3\times10^{-2}$m (red).}
    \label{fig:MULTIPLE_E}
\end{figure}
\begin{figure}
\begin{minipage}{0.60\linewidth}
    \centering
    \begin{tabular}{|c|c|c|c|c|}
    \hline
         Name & Young's modulus & $e_{mean}$ & $e_{max}$ & $\text{SNR}^{dB}$ \\
         \hline\hline
         Beam (a) & $9.6\times10^{10}$ & $3.9\times10^{-5}$ & $0.028$ & 25.66  \\
         Beam (b) & $7.6\times10^{9}$ & $4.8\times10^{-5}$ & $0.038$ & 25.42  \\
         Beam (c) & $1.0\times10^{11}$ & $4.1\times10^{-5}$ & $0.034$ & 24.25  \\
         Beam (d) & $6.4\times10^{10}$ & $6.8\times10^{-5}$ & $0.060$ & 23.73  \\
         \hline
    \end{tabular}
    \captionof{table}{Error values and SNR of the deformations shown at figure \ref{fig:MULTIPLE_E}. The displayed deformations are highly non-linear, yet the error values and SNR remain in the range of values displayed at table \ref{tab:Young_modulus_gen}.}
\end{minipage}
\hfill
\begin{minipage}{0.38\linewidth}
The error doesn't seem to have any specific pattern over few samples. In the long run, the pattern seen in the previous sections appears. All beams have similar errors in every metric with an average error around $5\times10^{-5}$ with a $SNR$ around of $24.5$. The only discrepancy appears with Beam (d) where the maximum error reaches $0.060$ and doubles the other values. The chosen Young's modulus doesn't seem to affect the accuracy of the network. 
\end{minipage}
\end{figure}

Beam (b) and Beam (c) are respectively near the lower and upper bound of the Young's modulus range and provide similar errors, both being comparable to the other two beams in term of errors too.

\subsubsection{ANN accelerated Newton-Raphson}\label{HNRA_results}
Artificial neural network has proven to be precise up to a couple of micro meters on average. Where one could be satisfied with the precision, another may want to ensure the respect of some properties such as incompressibility or fixed points on the boundaries. As shown in section \ref{HNRA}, the Hybrid Newton-Raphson algorithm proposes to use the prediction of the network in order to speed-up the algorithm.
The experiment will compare the speed and solution of the ANN, the classic Newton-Raphson and its hybrid version. The network trained for the beam at section \ref{gen_res} is used to obtain the following values.

\begin{table}[h]
    \centering
    \begin{tabular}{|c|c|c|c|}
    \hline
         Solver type & Converged simulations & Network prediction picked & Average iterations count\\
         \hline\hline
         Classic Newton-Raphson & $46 \%$ & - & 6.1\\
         Hybrid Newton-Raphson & $71 \%$ & $66 \%$ & 5.0\\
         \hline
    \end{tabular}
    \caption{Results of a comparison between the classic Newton-Raphson algorithm and the presented Hybrid Newton-Raphson algorithm over 100 randomly distributed forces with random amplitudes.}
\end{table}

These results are computed from a data set of 100 random external forces. Among them, 50 are within the amplitude range of the training data set and the 50 other have an amplitude up to two times bigger. Although 50 of them have an amplitude that is within the training bounds, they do not share orientation nor location with any of the training data set.

The classic Newton-Raphson manages to converge $46\%$ of the time. On average, when it converges, it does so in $6.1$ iterations.

The Hybrid Newton-Raphson converges $71\%$ of the time and $100\%$ of the time when the classic version did too. Our algorithm leads to a gain of $54\%$ in term of convergence. In $66\%$ of the cases, the solution of the neural network is preferred to the first Newton-Raphson iteration. Over the 100 test samples, the Hybrid Newton-Raphson picked two out of three times the prediction of the network as a better starting point than the result of the first iteration of the Newton-Raphson algorithm. From this point, on average, the algorithm converges in 5.0 iterations. This shows that from the prediction the algorithm converges on average in 4.0 iterations adding up to 5.0 to account for the first one that is discarded when the prediction is picked.
Overall the proposed algorithm converges more often and faster than the classic method while keeping the ordinary convergence properties of the Newton-Raphson algorithm.

\section{Discussion and Conclusion}\label{conclusion}
We presented a novel machine learning framework that can learn and compute very quickly and accurately complex nonlinear static deformations of elastic structures. This framework is based on modal analysis, a physics aware loss function and an optional Hybrid Newton-Raphson algorithm. Our results show that we can train a neural network with a small data set to achieve micrometer-scale errors in less than a millisecond, thus easily reaching real-time requirements. The method is also robust to Young's modulus generalisation, but requires more samples to achieve the same precision due to the increase in problem complexity. The Hybrid Newton-Raphson algorithm displays an important gain with a $54\%$ increase in convergence rate and $20\%$ gain in average iterations count.

The method shows promising results, but can be further improved. Firstly, some tests have shown that external forces applied very locally are not very well handled by the network and would need a more important representation in the data set. Secondly, the network has no knowledge of the topology, the geometry nor Dirichlet boundary conditions. Thus, any changes in these characteristics will not be taken into account in the predictions. This  limits the generalisation abilities of the network, although it remains similar to Model Order Reduction methods. Finally, for small displacements, the noise introduced by the network can be of the same amplitude as the range of motion. However, for small linear deformations, there is no need for any of the methods described in this paper, as more efficient and simpler solutions exist. 

In our approach, we rely on a fully connected network for its simplicity and efficiency. However, a change in the architecture of the artificial neural network could help with the generalisation and also reduce the number of parameters of the network. Architectures such as convolutional neural networks are known to generalise well from sparse data sets and we will consider them as a way to further develop our approach. In addition, recent developments in graph network can also alleviate the constraints on the topology typically imposed by CNNs, allowing to use any type of mesh rather than a regular grid structure.

%\SC{The modal analysis of the stiffness matrix to generate the data set could be used with other methods such as Model Order Reduction. This could greatly improve the guesses to generate interesting deformations and provide a sparser reduction thus better performances for the method.} \COM{$\longrightarrow$ je propose de ne pas commenter ce paragraphe et de rester uniquement sur le sujet des reseaux de neurones.}

\section{Acknowledgements}
This work was supported by the DRIVEN project grant agreement No 811099.

\newpage

\bibliography{references}

\end{document}